\pdfoutput=1

\documentclass[11pt]{article}

\usepackage[final]{acl}

\usepackage{times}
\usepackage{latexsym}
\usepackage{booktabs}

\usepackage{graphicx} 

\usepackage[T1]{fontenc}

\usepackage[utf8]{inputenc}

\usepackage{microtype}

\usepackage{inconsolata}

\usepackage{graphicx}

\usepackage{comment}
\usepackage{lipsum}
\usepackage{amsmath}
\usepackage{amsfonts}
\usepackage{graphicx} 
\usepackage{tablefootnote}

\usepackage{booktabs}
\usepackage{caption}
\usepackage{amsmath}
\usepackage{siunitx}
\usepackage{multirow}

\title{Query-based Cross-Modal Projector Bolstering Mamba Multimodal LLM}

\author{SooHwan Eom$^1$ \quad {\bf Jay Shim$^1$} \quad {\bf Gwanhyeong Koo$^1$} \quad {\bf Haebin Na$^1$} \\ {\bf Mark A. Hasegawa-Johnson$^2$} \quad {\bf Sungwoong Kim$^3$} \quad {\bf Chang D. Yoo$^{1}$}\Thanks{Corresponding author}
\\
  $^1$Korea Advanced Institute of Science and Technology / Korea, Republic of \\
  $^2$University of Illinois in Urbana-Champaign / United States of America \\
  $^3$ Korea University / Korea, Republic of
  \\
  $^1$\texttt{\{sean1105, shimjay17, kookie, sunbean0511, cd\_yoo\}@kaist.ac.kr}\\
  $^2$\texttt{jhasegaw@illinois.edu}\\
  $^3$\texttt{swkim01@korea.ac.kr}
}

\begin{document}
\maketitle
\begin{abstract}
  

The Transformer's quadratic complexity with input length imposes an unsustainable computational load on large language models (LLMs). In contrast, the Selective Scan Structured State-Space Model, or Mamba, addresses this computational challenge effectively. This paper explores a query-based cross-modal projector designed to bolster Mamba's efficiency for vision-language modeling by compressing visual tokens based on input through the cross-attention mechanism. This innovative projector also removes the need for manually designing the 2D scan order of original image features when converting them into an input sequence for Mamba LLM. Experimental results across various vision-language understanding benchmarks show that the proposed cross-modal projector enhances Mamba-based multimodal LLMs, boosting both performance and throughput.
\end{abstract}

\section{Introduction}

Multimodal Large Language Models (MLLMs) aim to extend the capabilities of Large Language Models (LLMs) to various modalities, including text and images. By fusing visual information into the textual domain, MLLMs effectively leverage the powerful language generation and logical reasoning abilities of text-only pre-trained LLMs. This integration has demonstrated significant potential in solving real-world vision-language problems, with diverse applications such as visual question answering (VQA) and multimodal dialogue response generation.

The core element behind this advancement lies in the Transformer \cite{transformer}, an architecture defined by stacked layers of attention mechanisms capable of scaling up to over 100 billion parameters. Due to its capability and flexibility to capture long-term dependencies, the Transformer can better represent different modalities, serving as a foundational model for MLLMs. Unfortunately, the Transformer also inherits intrinsic bottlenecks due to its defining attention mechanism. The computational and memory complexities of self-attention increase quadratically with sequence length, imposing a limit on the input sequence length. Recent efforts have focused on extending the Transformer's context window to overcome this limitation, but the challenge of computational burden remains.

To address this issue, the state-space model (SSM) \cite{lssl, s4, s4d, h3} has been studied as an alternative architecture for efficiently capturing long-range dependencies. The SSM can be viewed as combining Convolutional Neural Networks (CNNs) and Recurrent Neural Networks (RNNs), enabling parallelizable training and fast inference. The latest advancement in SSMs is Mamba \cite{mamba}, which incorporates an input-dependent gating mechanism that enables selective scanning, along with a hardware-aware algorithm for efficient computation. Mamba matches or even surpasses the performance of advanced Transformers while achieving faster training and inference speeds, leading to applications in various domains, including image \cite{vim, vmamba}, speech \cite{dpmamba, spmamba}, and video processing \cite{videomamba}. 
The utilization of Mamba architecture for MLLM foundation models has been considered \cite{vlmamba, cobra} but not extensively explored. Moreover, there remains a limited understanding of the most effective methods for aligning visual information within the textual domain using Mamba.

\begin{figure*}
  \centering
  \includegraphics[width=0.96\linewidth]{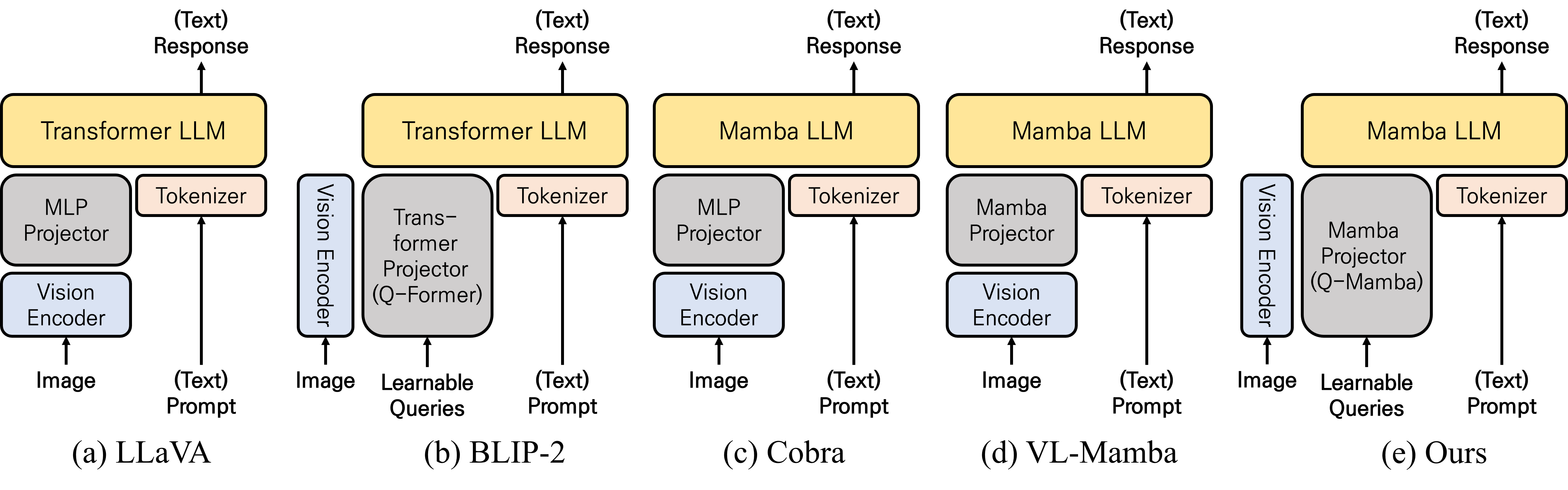}
  \caption{Model comparison between (a) LLaVA \cite{llava}, (b) BLIP-2 \cite{blip2}, (c) Cobra \cite{cobra}, (d) VL-Mamba \cite{vlmamba}, and (e) ours. The key differences stem from the choice of LLM backbone architecture, the design of the projector architecture, and the incorporation of learnable queries for flexibility.}
  \label{fig.comparison}
\end{figure*}

Building upon the previous architecture, we introduce a non-trivial Mamba-based architecture for cross-modal projection to connect the pre-trained vision encoder and Mamba-based LLM. Inspired by Querying Transformer (Q-Former) \cite{blip2}, we utilize learnable queries to project vision information from image features into 1D causal tokens by interleaving the Mamba sequence modeling layer and cross-modal attention. Our architectural design is motivated by three key objectives: (1) eliminating the heuristic choice of 2D visual scan order, (2) effectively and dynamically downsampling the projected visual feature sequence length, and (3) enhancing text-image alignment by adopting a structure tailored for Mamba-based multimodal modeling. We further propose MLLM with a pre-trained Mamba LLM backbone connected to the vision encoder using the proposed projector. The overall comparison between the previous models and ours is depicted in Figure \ref{fig.comparison}.

Our contributions can be summarized as follows:

\begin{itemize}
    \item We propose Querying Mamba, the multimodal connector based on the Mamba module, and the cross-modal attention for adaptive flexibility in downsampling the visual token lengths.
    
    \item We propose MLLM based on Querying Mamba and pre-trained Mamba LLM. We meticulously explore a range of choices regarding the components that integrate these models to boost Mamba's effectiveness in multimodal modeling.
    
    \item We carry out comprehensive experimental evaluations using multimodal comprehension benchmarks to assess the performance and robustness of our proposed models.
\end{itemize}

\section{Related Works}

\subsection{State-Space Models (SSMs) and Mamba}

Current state-space models are inspired by classical state-space models, which represent continuous systems that map a 1-dimensional function or sequence through an implicit latent state. The Linear State Space Layer (LSSL) \cite{lssl} was one of the earliest attempts at deep SSMs, aiming to enhance sequence modeling performance by stacking multiple SSM layers. Although LSSL demonstrated the potential of deep SSMs for addressing long-range dependencies, its high computational and memory costs rendered it impractical.

The Structured State-Space Model (S4) \cite{s4} tackled this bottleneck by re-parameterizing the latent matrix through decomposition into low-rank and normal terms. This innovation led to several variant architectures, such as the Diagonalized State-Space (DSS) \cite{dss} and S4D \cite{s4d}, which enabled more efficient and simplified computation via diagonalization. However, S4 and its variants can not remember specific past tokens or compare tokens across the sequence—capabilities crucial for language modeling. Hungry Hungry Hippos (H3) \cite{h3} aimed to overcome these shortcomings of S4 by incorporating 1-dimensional convolution along the sequence, allowing SSMs to compare and remember past tokens by shifting the input sequence.

The latest work, Mamba \cite{mamba}, further refines S4 by introducing a selective mechanism that utilizes input-dependent latent state parameters, making the model content-aware and enabling it to selectively focus on relevant information. Mamba also incorporates 1-dimensional convolution shifting from H3 and a gating mechanism similar to Long Short-Term Memory (LSTM) \cite{lstm}, which enhances its ability to handle long sequences with increased robustness and flexibility. With parallel associative scanning and a hardware-aware implementation, Mamba achieves efficient training and inference, matching or surpassing the capabilities of advanced Transformers.

The success of Mamba has led to various adaptations across different domains. For instance, several attempts have been made to apply Mamba in speech separation \cite{spmamba, dpmamba}. In computer vision, Vision Mamba (Vim) \cite{vim} and V-Mamba \cite{vmamba} employ bidirectional SSMs to process two-dimensional image data with one-dimensional sequence modeling in Mamba. SiMBA \cite{simba} further enhances this by incorporating a channel-mixing layer into the Mamba block, analogous to the role of the feedforward network in the Transformer block.

\subsection{Multimodal Large Language Models}

With the introduction of ChatGPT \cite{chatgpt}, also referred to as InstructGPT, Large Language Models (LLMs) have emerged as a dominant approach for real-world natural language processing tasks. These models, typically featuring billions of parameters and trained on extensive corpora, are not only proficient in generating language responses but also in tasks requiring logical comprehension and reasoning. Although InstructGPT has not been publicly released, the research community has been actively developing open-source LLMs \cite{llama, phi1, phi1.5, opt}, which have shown performance on par with InstructGPT. This progress has led to various adaptations and modifications of pre-trained LLMs for diverse applications.

A notable advancement is the development of Multimodal Large Language Models (MLLMs), which leverage pre-trained LLMs to process multimodal data. This extends beyond the original text-only domain, integrating capabilities to understand both textual and visual inputs. Models like LLaVA \cite{llava}, BLIP\cite{blip, blip2}, and GPT-4\cite{gpt4} have shown robust performance in tasks requiring nuanced vision-language integration. These models utilize transformer-based frameworks known for handling long-range dependencies effectively. However, the innate characteristic of high computational demands and slow inference rates of these transformer-based frameworks have started to become a target for recent research, leading to the adoption of the more efficient Mamba architecture in MLLMs. This initiative has given rise to models like Cobra\cite{cobra} and VL-Mamba\cite{vlmamba}, which demonstrate promising pathways for enhanced efficiency in MLLM deployment.


Cobra \cite{cobra} employs a state-space model for multimodal tasks, leveraging the linear scalability of the Mamba architecture. It introduces an innovative approach to vision encoding by merging outputs from DINOv2 \cite{dino2} and SigLIP \cite{siglip}, thereby generating visual representations that capture both spatial and semantic properties effectively. These outputs are then processed through a learnable projector module, which aligns the visual and textual features by adjusting the dimensions of the visual representations to match those of the Mamba LLM via a multi-layer perceptron. This approach enables Cobra to deliver the same volume of output tokens in just 30\% of the time required by comparable 3B transformer-based LLMs, such as TinyLLaVA \cite{tinyllava} or MobileVLM v2 \cite{mobilevlm2}.

Similarly, VL-Mamba \cite{vlmamba} builds upon a pretrained Mamba framework and introduces a novel MultiModal Connector (MMC) architecture. This connector features a Vision Selective Scan (VSS) module and two linear layers, which enhance the causal relationships among image blocks from the vision encoder. Furthermore, this paper assesses the performance difference between the Bidirectional-Scan Mechanism (BSM), which scans the image blocks in both forward and backward directions, and the Cross-Scan Mechanism (CSM), which scans both from forward to backward and top to bottom. This paper suggests a preference for the simple BSM, as the two scanning methods show comparable efficacy.

However, the previous projector modules used in Cobra and VL-Mamba have limitations in that these connectors have no flexibility in vision token number, causing longer vision token input, and require manual scan mechanisms that grant causality between image blocks. 

\begin{figure*}
  \centering
  \includegraphics[width=0.98\linewidth]{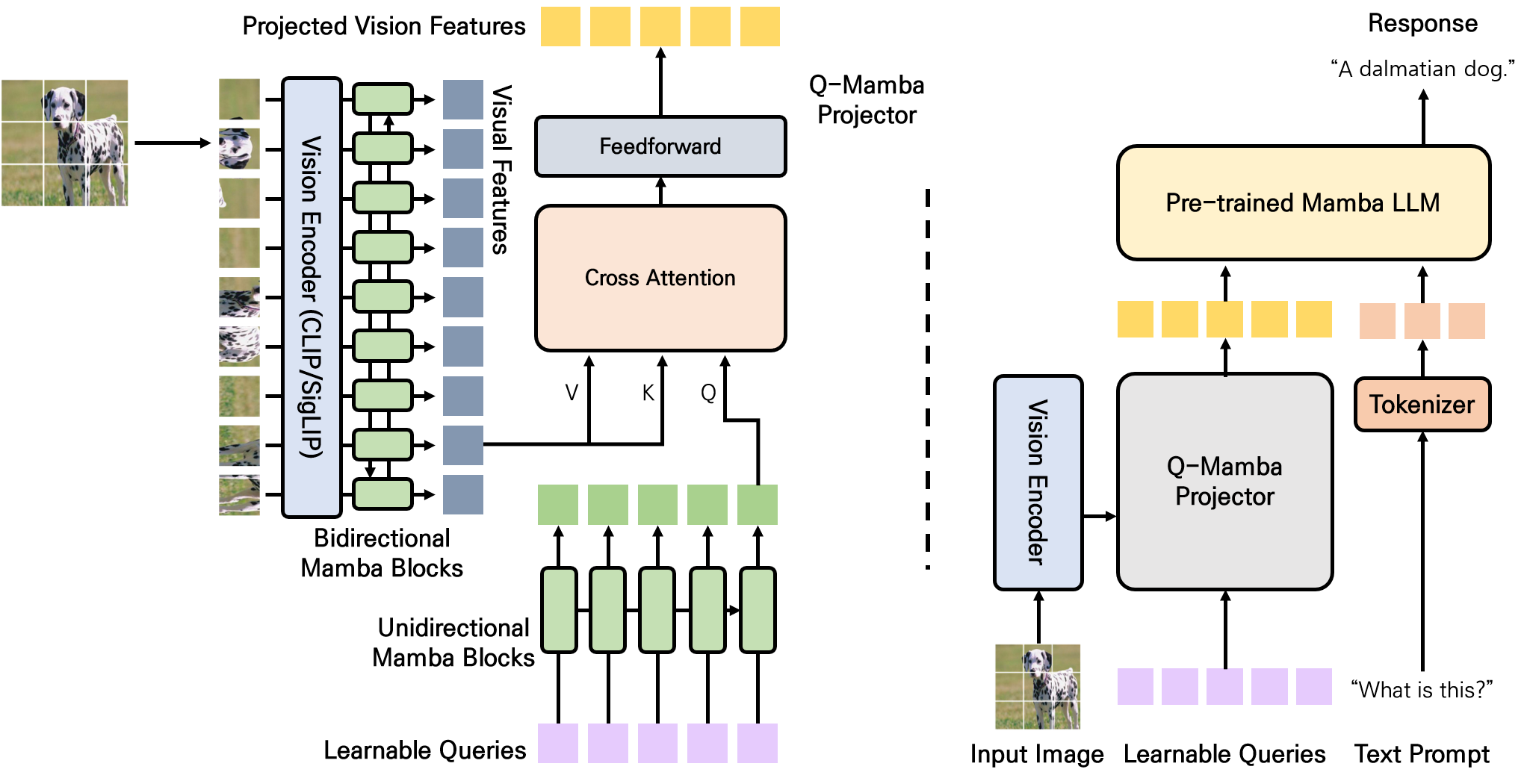}
  \caption{Overall architecture of Querying Mamba (left) and the Multimodal Mamba LLM (right) based on the proposed design. Querying Mamba projects the visual information, which is encoded by a pre-trained vision encoder with an additional bidirectional Mamba layer, into the learnable queries with causal Mamba prior via cross attention. The projected vision features work as vision token inputs for pre-trained Mamba LLM.}
  \label{fig.qmamba}
\end{figure*}

\section{Method}

In this section, we first review the preliminary concepts of structured state-space models and Mamba (Sec. \ref{prelim}). Then, we describe the details of the Cross-modal Mamba projector, which extracts the 2-dimensional vision information into a 1-dimensional causal token sequence (Sec. \ref{qmamba}). Lastly, we describe the two-stage fine-tuning of the multimodal Mamba with our proposed Q-Mamba (Sec. \ref{mllm}).

\subsection{Preliminaries} \label{prelim}

State-Space Models (SSMs) \cite{lssl, s4, s5} represent linear time-invariant systems that map a continuous 1-dimensional function or a sequence $x(t) \in \mathbb{R}$ to a corresponding response $y(t) \in \mathbb{R}$, via a hidden state $h(t) \in \mathbb{R}^N$ with $N$ latent dimensions. These systems are characterized by four parameters $(\mathbf{A}, \mathbf{B}, \mathbf{C}, \mathbf{D})$, which define the system dynamics and outputs as follows:
\begin{equation}
\begin{split}
    h'(t) &= \mathbf{A} h(t) + \mathbf{B} x(t) \\
    y(t) &= \mathbf{C} h(t) + \mathbf{D} x(t)
\end{split}
\end{equation}
Typically, the parameter $\mathbf{D}$ is omitted as it can be interpreted as a skip connection, which is computationally straightforward to implement.

In practice, to deal with discrete-time input sequences, SSMs are discretized with matrices $\overline{\mathbf{A}}$ and $\overline{\mathbf{B}}$. One common discretization method is the Zero-Order Hold (ZOH) method, outlined as:
\begin{equation}
\begin{split}
\overline{\mathbf{A}} &= \exp (\mathbf{\Delta}\mathbf{A}) \\
    \overline{\mathbf{B}} &= (\mathbf{\Delta}\mathbf{A})^{-1} (\exp (\mathbf{\Delta}\mathbf{A}) - \mathbf{I}) \cdot (\mathbf{\Delta}\mathbf{B})
\end{split}
\end{equation}
where the parameter $\mathbf{\Delta}$ specifies the discretization step size. The reformulated discretized system is given by:
\begin{equation}
\begin{split}
    h_t &= \overline{\mathbf{A}} h_{t-1} + \overline{\mathbf{B}} x_t \\
    y_t &= \mathbf{C} h_t
\end{split}
\end{equation}

Structured State-Space Model (S4) \cite{s4} operates as a time-invariant system, meaning its defining parameters $(\mathbf{A}, \mathbf{B}, \mathbf{C}, \mathbf{\Delta})$ remain constant across all time-steps. Mamba \cite{mamba} addresses this constraint by making $\mathbf{B}$, $\mathbf{C}$, and $\mathbf{\Delta}$ input-dependent, enabling a dynamic gating mechanism based on the input sequence. This allows Mamba to selectively focus on pertinent information, significantly enhancing its language modeling capabilities.

\subsection{Cross-Modal Mamba Projector} \label{qmamba}

\begin{figure}
  \centering
  \includegraphics[width=\linewidth]{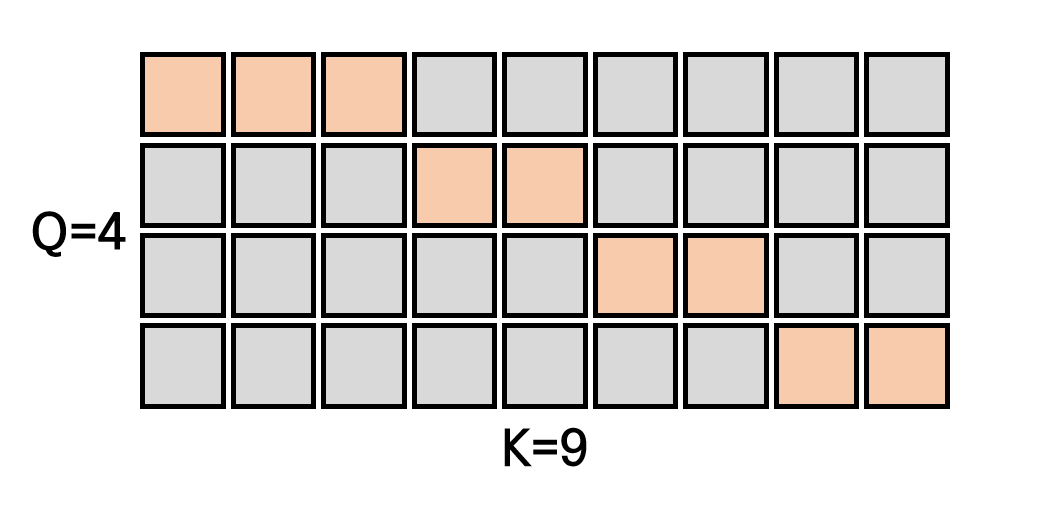}
  \caption{Example of local attention mask applied in the cross-attention layer inside Querying Mamba with 4 queries ($Q$) and 9 keys ($K$). Each query attends exclusively to $K/Q$ keys, enabling the focused extraction of information from distinct visual components.}
  \label{fig.local}
\end{figure}

We introduce the cross-modal projector, Q-Mamba, which integrates the Mamba architecture with cross-attention. The architecture of Q-Mamba, illustrated on the left side of Figure \ref{fig.qmamba}, comprises stacked Q-Mamba blocks, each containing a Mamba layer, cross-attention, and a feedforward network. The Mamba layer functions as a sequence mixer, while the feedforward network acts as a channel mixer. A set of learnable query embeddings serves as the input sequence to Q-Mamba. This unidirectional Mamba layer introduces causal dependencies between the queries, ensuring a structure that enhances compatibility with the LLM's sequential processing. These queries interact with vision features from the frozen pre-trained vision encoder via cross-attention layers, enabling access to arbitrarily positioned encoded visual information. We found that applying a local attention mask, as shown in Figure \ref{fig.local}, empirically improves model performance.

This design aligns the projected visual features with the language understanding capabilities of the pre-trained LLM, facilitating seamless integration of visual and textual input. The causal prior further enhances this alignment, ensuring that query embeddings are coherent and compatible with the LLM’s sequential processing.

Q-Mamba offers three key advantages for cross-modal projection. First, it is independent of visual scan order. Previous Mamba-based vision encoders relied on heuristic scan order choices, such as bidirectional or cross-directional scans \cite{vlmamba, vim, vmamba}. Q-Mamba removes this reliance by using cross-attention to project visual information from arbitrarily ordered image features onto a one-dimensional query sequence. Second, the model allows flexibility in choosing the query sequence length. Direct application of Mamba on visual feature sequences often results in projected features of equivalent length, which can be excessive for Mamba LLM. Q-Mamba, however, enables effective downsampling of the visual feature length. Finally, the architecture’s similarity to Q-Former \cite{blip2} from transformer-based MLLMs ensures proper alignment of text-image features.

We explore several architectural variants to identify the optimal configuration for Q-Mamba. Our investigation includes the use of bidirectional Mamba for preprocessing visual features, the incorporation of a feedforward network for channel mixing, and determining the optimal length of the learnable query sequence. The findings are detailed in Section \ref{ablations}.

\subsection{Multimodal Mamba Language Model} \label{mllm}

We introduce the MLLM based on our querying cross-modal projector (Q-Mamba). As shown in Figure \ref{qmamba}, the overall architecture consists of a pre-trained vision encoder, our cross-modal projector, and a pre-trained Mamba LLM. Initially, visual features are extracted from the input image using the vision encoder. These features are then processed by our projector, which outputs queries embedded with projected visual information. Subsequently, this output sequence is combined with a tokenized text prompt and fed into the Mamba LLM, which generates the corresponding text response.

\paragraph{Training}

We adopt a two-stage training scheme from LLaVA \cite{llava}, where the initial stage involves aligning the projected features within the frozen LLM using a filtered visual instruction-following dataset. The subsequent stage entails end-to-end fine-tuning of both the projector and the LLM using an extensive visual instruction-following dataset.

\begin{table*}[h!]
\centering
\resizebox{\textwidth}{!}{
\begin{tabular}{l c | c c c c c c | c}
\toprule
\textbf{Name} & \textbf{Query Length} & \textbf{VQA\textsuperscript{v2}} & \textbf{GQA} & \textbf{VizWiz} & \textbf{VQA\textsuperscript{T}} & \textbf{POPE} & \textbf{MMB} & \textbf{sec/iter (Training / Inference)} \\
\midrule
Cobra$^*$ & - & 75.38 & 58.16 & 49.22 & 44.9 & 87.6 & 56.2 & 7.62 / 0.129 \\
\midrule
VL-Mamba$^*$ & - & 74.38 & 56.69 & \textbf{51.66} & 48.7 & 83.9 & 57.0 & 7.26 / 0.152 \\
+ forward scan only & - & 72.34 & 51.92 & 29.17 & 45.6 & 85.9 & 56.7 & -\\
+ backward scan only & - & 72.06 & 52.42 & 34.92 & 45.1 & 86.1 & 55.9 & -\\
\midrule
Ours & 128 & 74.51 & 57.59 & 51.03 & 47.1 & \textbf{87.9} & 57.2 &  \textbf{5.52} / \textbf{0.095} \\
Ours & 256 & 75.01 & 58.10 & 50.53 & 48.8 & 86.9 & 57.7 & 5.94 / 0.099\\
Ours & 512 & 75.42 & \textbf{58.37} & 48.90 & 50.2 & 86.5 & 57.6 & 6.54 / 0.127\\
Ours & 729 & \textbf{75.62} & 58.33 & 49.30 & \textbf{51.2} & 86.8 & \textbf{58.0} & 7.54 / 0.147\\
\bottomrule
\end{tabular}
}
\caption{Comparison with Multimodal Mamba LLMs on 6 benchmarks: VQA\textsuperscript{v2} \cite{vqav2}, GQA \cite{gqa}, VizWiz \cite{vizwiz}, VQA\textsuperscript{T} (TextVQA) \cite{vqat}, POPE \cite{pope}, and MMB (MMBench) \cite{mmbench}. $^*$ indicates the results were reproduced within the same codebase and experimental conditions for fair comparison. We also examined variants of the previous Multimodal Mamba LLMs: + forward scan only and + backward scan only indicate the visual scanning order of multimodal connector inside VL-Mamba \cite{vlmamba}. We also report the time consumed per fine-tuning and inference iteration in seconds.}
\label{table:model_evaluation}
\end{table*}

\section{Experiments}

\subsection{Settings}

\paragraph{Datasets}

For the fine-tuning stage, we follow the existing two-stage training paradigm and dataset based on LLaVA \cite{llava} with additional datasets. For the alignment stage, we use a filtered dataset from CC3M with 595K image-text pairs. For the end-to-end fine-tuning stage, we use the combined dataset consisting of LLaVA v1.5 mixed dataset \cite{llava} with 655K visual conversations, LVIS-Instruct-4V \cite{lvis} dataset with 220K context-aware visual instruction pairs, and LRV-Instruct dataset \cite{lrv} with 400K visual instruction pairs aimed for hallucination mitigation.  

\paragraph{Models}

For the pre-trained vision encoder, we employ pre-trained SigLIP \cite{siglip}, which encodes vision features for each patched image. We utilize a ViT structure with 400 million parameters. The input image resolution is configured at $384 \times 384$, and the total number of visual features is $729$. We also attached a bidirectional multimodal connector from trained VL-Mamba \cite{vlmamba} to the vision encoder. The output of the multimodal connector is used as a vision feature input for the Q-Mamba projector.

The backbone of our model is the pre-trained Mamba \cite{mamba} LLM, which consists of 2.8 billion parameters. This model was initially pre-trained on the SlimPajama datasets \cite{slimpj} for 600 billion tokens, instruction-tuned on the UltraChat 200K dataset \cite{ultrachat}, and then fine-tuned on the UltraFeedback dataset \cite{ultrafeedback} using Direct Preference Optimization (DPO) \cite{dpo}.

For the Q-Mamba projector, we stack 24 blocks with an inner dimension of 768. This choice of hyperparameter is to copy the pre-trained weights of Mamba \cite{mamba} with the size of 130M parameters.

\paragraph{Training}

We train the model using four NVIDIA A100 80GB GPUs. During training, we leverage the PyTorch Fully Sharded Data Parallel \cite{fsdp} framework, utilizing automatic mixed-precision with FP32 and BF16 for efficient distributed training. The batch sizes are set to 256 for the alignment stage and 128 for the end-to-end fine-tuning stage. We employ the Rectified Adam (RAdam) optimizer \cite{radam}, coupled with a cosine decay learning rate scheduler. The learning rates are set at $1\times10^{-4}$ for the alignment stage and $2\times10^{-5}$ for the end-to-end fine-tuning, both with a warmup ratio of $0.03$. Each training stage is conducted in a single epoch.


\begin{table*}[h!]
\centering
\begin{tabular}{c | c c c c c c}
\toprule
\textbf{Attention} & \textbf{VQA\textsuperscript{v2}} & \textbf{GQA} & \textbf{VizWiz} & \textbf{VQA\textsuperscript{T}} & \textbf{POPE} \\
\midrule
Global & 73.12 & 52.87 & 49.09 & 44.0 & 85.1  \\
\midrule
Local & 75.01 & 58.10 & 50.53 & 48.8 & 86.9 \\
\bottomrule
\end{tabular}
\caption{Comparison between global attention and local attention for cross-attention layer inside our cross-modal Mamba projector. We used 256 learned queries for both models.}
\label{table:global_vs_local}
\end{table*}

\begin{table*}[h!]
\centering
\begin{tabular}{c | c c c c c c}
\toprule
\textbf{Bi-directional Mamba} & \textbf{VQA\textsuperscript{v2}} & \textbf{GQA} & \textbf{VizWiz} & \textbf{VQA\textsuperscript{T}} & \textbf{POPE} \\
\midrule
From Scratch & 74.22 & 56.30 & 53.12 & 48.0 & 86.4  \\
\midrule
From Trained & 75.01 & 58.10 & 50.53 & 48.8 & 86.9 \\
\bottomrule
\end{tabular}
\caption{Comparison between using bidirectional multimodal connector inside vision encoder from scratch or from trained VL-Mamba \cite{vlmamba}. We used 256 learned queries and local attention for both models.}
\label{table:scratch_vs_trained}
\end{table*}

\begin{table*}[h!]
\centering
\begin{tabular}{c | c c c c c c}
\toprule
\textbf{Visual Scan Order} & \textbf{VQA\textsuperscript{v2}} & \textbf{GQA} & \textbf{VizWiz} & \textbf{VQA\textsuperscript{T}} & \textbf{POPE} \\
\midrule
Forward Only & 76.58 & 58.44 & 50.00 & 50.0 & 86.9  \\
Backward Only & 75.35 & 58.13 & 49.51 & 50.4 & 86.7 \\
\midrule
Bidirectional & 75.01 & 58.10 & 50.53 & 48.8 & 86.9 \\
\bottomrule
\end{tabular}
\caption{Comparison between using raster scan only or bidirectional multimodal connector inside vision encoder from trained VL-Mamba \cite{vlmamba} during inference-time. We used 729 learned queries and local attention for both models.}
\label{table:forward_vs_bidirectional}
\end{table*}

\paragraph{Evaluation}

To validate the performance of our model, we benchmarked it against five different datasets: VQA-v2 \cite{vqav2}, GQA \cite{gqa}, VizWiz \cite{vizwiz}, Text-VQA \cite{vqat}, POPE \cite{pope} and MMBench \cite{mmbench}. Each dataset offers unique challenges and measures different aspects of the model's capabilities:
\begin{itemize}
    \item VQA-v2  \cite{vqav2} evaluates the model's general ability to reason over Vision-Question pairs.
    \item GQA \cite{gqa} extends VQA-v2 by testing the model's reasoning skills across a broader spectrum, incorporating spatial understanding and multi-step inference along with various reasoning skills.
    \item VizWiz \cite{vizwiz}, similar to VQA-v2, includes unanswerable questions, thereby assessing the model's ability to identify when a question cannot be answered. 
    \item Text-VQA \cite{vqat} specifically measures the model's proficiency in recognizing text within images and answering related questions. 
    \item POPE \cite{pope} differentiates itself by focusing on the model's susceptibility to hallucination problems. It provides a score based on the probability of the given answer, hence evaluating the likelihood that the model avoids generating incorrect information. 
    \item MMBench \cite{mmbench}  evaluates the multi-modal capabilities of vision-language models across 20 distinct abilities, including object localization, social reasoning, and fine-grained perception. It introduces a novel CircularEval strategy, ensuring comprehensive evaluation through multiple passes of QA to reduce biases and improve reliability.
\end{itemize}

\subsection{Results}
As presented in Table 1, our model consistently outperforms previous state-of-the-art Mamba-based multimodal models across all benchmarks. Specifically, the Q-Mamba with 729 queries achieves the highest overall performance, demonstrating significant improvements in tasks that require nuanced vision-language integration.

The results in Table 1 indicate that increasing the number of queries generally improves performance. For instance, moving from 128 to 256 queries results in substantial performance gains across all benchmarks, highlighting the importance of having a sufficient number of queries to capture detailed visual information. Further increasing the number of queries to 512 and 729 continues to improve performance, though the gains are less pronounced compared to the initial increase. However, further increases to 512 and 729 queries show diminishing returns, as additional queries yield progressively smaller benefits.

Metrics such as VizWiz and POPE, which evaluate the model's ability to identify unanswerable questions and assess hallucination risk respectively, exhibit an inverse relationship with query size. Although larger query sizes can capture more detailed visual information, they also tend to introduce extraneous data. This surplus of information can complicate the decision-making process in certain tasks, where the model is required to distinguish relevant from irrelevant details. As a result, slight performance drops are observed for the tasks mentioned, where excessive data may hinder the model’s accuracy.

Furthermore, Q-Mamba’s architecture significantly enhances throughput by dynamically downsampling visual feature sequences into compact semantic tokens, thereby reducing the computational burden on the LLM backbone. This streamlined design, coupled with flexible query sequence lengths, allows Q-Mamba to achieve an optimal balance between computational efficiency and performance. Such adaptability makes Q-Mamba highly suitable for diverse applications that demand both processing speed and accuracy.

\subsection{Ablation Studies} \label{ablations}
In our ablation study, we meticulously analyzed various configurations to determine how different components within Q-Mamba affect model performance. Our initial investigations focused on the type of cross-attention mechanism employed, with results detailed in Table \ref{table:global_vs_local}. These findings demonstrate that local attention significantly outperforms global attention in enhancing model performance. We then evaluated the effect of utilizing pre-trained weights for the bidirectional Mamba connector within the vision encoder, with outcomes presented in Table \ref{table:scratch_vs_trained}. The results confirm that leveraging weights from a trained VL-Mamba model leads to performance improvements. Finally, we explored the influence of the visual scan order in the bidirectional Mamba connector, as shown in Table \ref{table:forward_vs_bidirectional}. Interestingly, our data indicate that although the model is trained with a bidirectional scan setting, employing only a forward Mamba for inference does not decrease performance and can even enhance it.

\section{Conclusion}

This paper presents a query-based cross-modal projector designed to enhance Mamba's efficiency in multimodal vision-language modeling. By using the cross-attention mechanism between the learnable queries and the outputs of the visual encoder within a Mamba architecture, the proposed multimodal projector dynamically compresses visual tokens based on an input image context, eliminating the need for manually designing of the 2D scan order of image features. Experimental results on diverse vision-language understanding benchmarks demonstrate that the proposed cross-modal projector boosts the effectiveness of Mamba-based MLLMs.

\section*{Acknowledgements}
This work was partly supported by Center for Applied Research in Artificial Intelligence (CARAI) grant funded by DAPA and ADD (UD230017TD) and partly supported by Institute of Information \& Communications Technology Planning \& Evaluation (IITP) grant funded by the Korea government(MSIT) (No. RS-2019-II190079, Artificial Intelligence Graduate School Program(Korea University)).

\section*{Limitations}
Despite the promising results, our approach has several limitations that need to be addressed in future work. The primary limitation is related to the amount and quality of the dataset used for training and fine-tuning the model.

For the alignment process, we used the LLaVA-LLVIS dataset, and for the fine-tuning process, we used the LLaVA-1.5 dataset. Both of these datasets are filtered and curated to ensure quality, but their limited size compared to the vast datasets typically used in training large language models (LLMs) can restrict the model's ability to generalize across diverse vision-language tasks. Specifically, we ran one epoch for each stage of our training process, whereas other models in the same domain were fine-tuned for two epochs instead of one. This difference in training duration can result in less robust model performance, as the additional epochs in other models allow for more comprehensive learning and fine-tuning of the parameters.

Additionally, the Mamba architecture is liable to "forget." The hidden states of the Mamba model take input and output sequentially, similar to how hidden states within the RNN would, where the current state depends on the previous inputs and hidden state outputs. This sequential dependency can potentially result in forgetting issues that plagued the RNN/LSTM-based models, for long input.

It would be necessary to pretrain the proposed Q-Mamba more extensively including contrastive learning as used in Q-Former based on image-text pair datasets. In addition, the parameters of Q-Mamba can be initialized by the pre-trained compact Mamba LLM. Also, it would be helpful to perform a more in-depth analysis of the resulting attention map for each query according to different input images.

\section*{Potential Risk}
This paper presents a new architecture of a Large Language Model with over a billion parameters, which can cause potential discrimination in the use of these methods due to the disparity in access to computational resources. Also, the hallucination of the Large Language Model can cause potential bias or harm when generating the response.

\bibliography{ref}

\appendix



\end{document}